\documentclass[runningheads]{llncs}
\usepackage[T1]{fontenc}
\usepackage{amsmath,amsfonts,bm}
\usepackage{graphicx}
\usepackage{subfig}
\usepackage{wrapfig}
\usepackage{caption}
\usepackage{color}
\usepackage{soul}

\usepackage[utf8]{inputenc}
\usepackage{listings}

\usepackage{color}

\newcommand{\R}{\mathbb{R}}

\newcommand{\cD}{\mathcal{D}}

\newcommand{\Span}{\mathrm{span}}

\begin{document}

\title{Bioinspired CNNs for border completion in occluded images} 
\titlerunning{Bioinspired CNNs}

\author{Catarina P. Coutinho\inst{1}\orcidID{0000-0003-3707-8795} \and
Aneeqa Merhab\inst{2}\orcidID{0009-0004-8013-3061} \and
Janko Petkovic\inst{3}\orcidID{0009-0009-7455-9484} \and
Ferdinando Zanchetta \inst{1}\orcidID{0000-0003-2294-2755}  \and
Rita Fioresi \inst{1}\orcidID{0000-0003-4075-7641}
}
\authorrunning{Coutinho et al.}
%
\institute{Department of Pharmacy and Biotechnology, University of Bologna, Via San Donato 15, 40127 Bologna, Italy
\email{catarina.praefke2@unibo.it,}
\email{rita.fioresi@unibo.it,}
\email{ferdinando.zanchett2@unibo.it} \and
Department of Mathematics, University of Ferrara, Via Ariosto 35, 44122 Ferrara, Italy
\email{aneeqa.mehrab@unife.it} \and
Numerical Analysis and Applied Mathematics, VIB - KU Leuven, ON5/B Herestraat 49 - Bus 4039, 3000 Leuven, Belgium
\email{janko.petkovic@kuleuven.be}
}
\maketitle              
\begin{abstract}
  We exploit the mathematical modeling of the border completion problem in the visual cortex to design convolutional neural network (CNN) filters that enhance robustness to image occlusions. We evaluate our CNN architecture, BorderNet, on three occluded datasets—MNIST, Fashion-MNIST, and EMNIST—under two types of occlusions: stripes and grids. In all cases, BorderNet demonstrates improved performance, with gains varying depending on the severity of the occlusions and the dataset.
\end{abstract}

\section{Introduction}

\begin{wrapfigure}{r}{40mm}
  \begin{center}
    \includegraphics[scale=0.4]{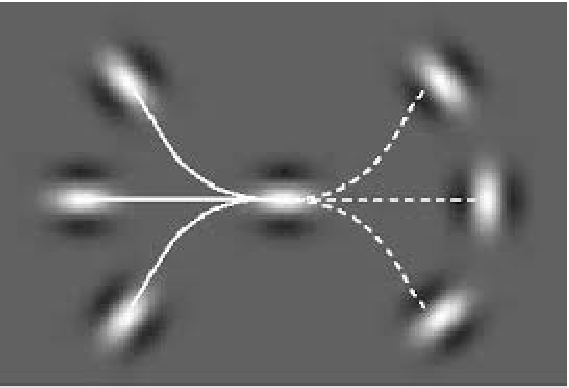}
  \end{center}
  \caption{Association Fields}\label{Hayes}
\end{wrapfigure}

Visual encoding of information in higher mammals faces significant challenges, as for example, situations where
occlusions partially hide an object, making its identification nontrivial. The mathematical modelling of mammalian visual cortex (V1) ability to reconstruct partially occluded images, is a current and active research area (see \cite{Citti2014NeuromathematicsOV} and refs. therein). Starting from the pioneering study by Hubel and Wiesel,
\cite{Hubel1969AnatomicalDO,Hubel1974UniformityOM}, whose ice-cube model of the visual cortex {first provided a functionally meaningful anatomic characterization of the primary visual cortex}, later models focused in understanding finer mechanisms as in \cite{Hoffman1989TheVC}, where Hoffman provides a mathematical representation of V1 as a contact bundle, explaining the orientation selectivity occurring in contour detection. In fact, anatomically, the border directions in a perceived image are selected within hypercolumns, the functional units in V1 composed of columns of neurons tuned to specific orientations. Each column responds maximally to a preferred orientation, enabling the construction of a local orientation map across the visual field. Information from the dominant orientation at each location is then relayed to higher cortical areas for further processing \cite{field}. A key functional property emerging from this hypercolumnar organization is {\sl contour integration}, which is the ability to reconstruct partially occluded boundaries from their visible portions. This phenomenon was first described on a phenomenological level by Gestalt psychology under the law of good continuation, one of the Prägnanz principles governing perceptual organization \cite{Mather2006FoundationsOP}. Continuity plays a central role in
shape perception, allowing the visual system to integrate spatially distributed orientation cues into smooth and coherent contours. Neurophysiologically, this capability is supported by the horizontal connectivity between hypercolumns: neurons with collinear orientation preferences tend to excite each other, while orthogonally tuned neurons are inhibited. These association fields as in Figure \ref{Hayes}, \cite{field} effectively promote the completion of interrupted contours, providing a biological solution to the occlusion problem.

This understanding of receptive fields of simple and complex cells, together with the anatomically observed structured arrangement of the orientation selective columns, led to finer mathematical formulations grounded in
differential geometry (see \cite{bc}, \cite{Bressloff2002TheVC}, \cite{Petitot1999VersUN}, \cite{citti}, \cite{alekseevski}). In particular, modeling V1 as a contact bundle as in \cite{Hoffman1989TheVC} led to take
advantage of the natural sub-Riemannian structure thus providing a rigorous framework in which contour completion can be interpreted as the solution of a sub-Riemannian geodesic problem. At the same time a growing number of
researchers in machine learning, started to draw a closer parallellism between the human visual perception and the successful algorithms of deep learning (see \cite{Fukushima1980NeocognitronAS}, \cite{Ecker2018ARC},
\cite{Duits2010LeftinvariantPE} and refs. therein).

In this work, we want to present the border completion in V1 from a theoretical point of view, via the Hamiltonian sub Riemannian formalism along the lines in \cite{bike}, which however had a different scope. Then, we shall take advantage of the mathematical modelling to investigate how the biological mechanisms underlying contour integration can be translated into a computational framework to enhance convolutional neural networks (CNNs) robustness against
occlusions in image classification tasks. Building on our previous geometric models of V1, we introduce biologically inspired directional filters into LeNet5, a CNN whose architecture bears notable similarities to the early visual pathway \cite{jf}. These predefined filters mimic the action of orientation-selective receptive fields and border integration operators.

We test our proposed architecture BorderNet on occluded versions of the MNIST, FashionMNIST and EMNIST datasets, where images are corrupted by diagonal black stripes, and grids. Our model is trained exclusively on the original,
unmodified dataset; occluded images are used only at test time. Our results show that incorporating orientation-aware filters significantly improves classification accuracy under such occlusions, thus confirming the proof of concept preliminary results obtained in \cite{Coutinho2025EnhancingCR}.

This paper is organized as follows.

Section 2 reviews the mathematical framework underlying our approach and details the proposed bio-inspired architecture and dataset construction.

Sections 3 and 4 present and discuss the experimental results, comparing the performance of the standard LeNet5 model with that of the enhanced network endowed with border integration capabilities.

\section{Mathematical Modeling for border completion in V1}

In this section we discuss the mathematical modeling for the mechanism of border completion in the primary visual cortex V1, via the calculation of the sub Riemannian geodesics with respect to a bracket generating distribution.
We shall use the Hamiltonian formalism, instead of the more established Lagrangian one \cite{citti}, \cite{Petitot1999VersUN}. These calculations appeared originally in \cite{bike}; we briefly recap them, since the different focus and scope of \cite{bike} may not make the reader aware of their significance for the present purpose.

\medskip
In sub Riemannian geometry, the metric on a manifold $M$ is replaced by a sub Riemannian metric, which is a semidefinite positive symmetric bilinear form $g$ on the tangent bundle, typically assumed to be of
constant (non maximal) rank (see \cite{Agrachev2019ACI}, \cite{Montgomery2006ATO} and refs. therein).
Hence its non degenerate locus defines a distribution $\cD$ on $M$, i.e. a smooth assignment
\begin{equation}\label{Ddef} 
M \ni x \mapsto \cD_x \subset T_xM, \qquad g|_{\cD_x} \, \hbox{non degenerate,}
\end{equation}
where dim $\cD_x < $ dim $T_xM$.
We call a distribution  
\textit{bracket generating}

when a finite number of brackets of the vector fields that generate it, generate the tangent space to $M$ at each point. Typically (but not always) in sub Riemannian geometry we assume the distribution $\cD$ as in (\ref{Ddef}) to be bracket generating. We call \textit{sub Riemannian geodesics} for the sub Riemannian metric $g$ the shortest paths
connecting two given points in the manifold $M$. In general, the sub Riemannian geodesic problem may not have a solution and when the solution exists it may not be unique: the Hamiltonian or Lagrangian formulation of the question
may indeed lead to different solutions \cite{Montgomery2006ATO}. However, despite the fact the two approaches are not in general equivalent in sub Riemannian geometry, with some extra regularity hypotheses, as for example
the assumption of $\cD$ coming from a principal bundle distribution, they are. We will see that indeed, viewing
the visual cortex as contact bundle as in \cite{Hoffman1989TheVC}, gives to it a natural a principal bundle structure \cite{alekseevski}, so that the Hamiltonian and Lagrangian formulations for the geodesic question lead to the same answer.

\medskip
We start by modeling the functional architecture of the primary visual cortex in its border completion capability,
(see \cite{Petitot1999VersUN}, \cite{citti}, \cite{Citti2014NeuromathematicsOV} and refs therein). We represent an image by a smooth function $I:D\subset\R^2\to\R$, each point $(x,y) \in D$ with $I(x,y)$ giving its grey level.

We define the vector field $Z$ 
on the manifold $\mathcal{E}=\R^2 \times S^1$ with
coordinates $(x,y,\theta)$:
\[
Z = -\sin\theta\,\partial_x + \cos\theta\,\partial_y .
\]
where $(x,y)$ are the coordinate of a point in $\R^2$ and $\theta$ represents an angle.

The vector field $Z$ is orthogonal to the line with angle $\theta$ with the $x$-axis at each point.

\medskip
Motivated by the action of orientation-selective cells, in the visual cortex, following \cite{jf}, \cite{bike}, we give a key definition.

\begin{definition}\label{ormap}
{\rm Let $\mathrm{reg}(D)$ be the set of points where $\nabla I\neq 0$.  
The \emph{orientation map} of $I$ is
\[
\Theta(x,y) = \operatorname*{argmax}_{\theta\in S^1}
\big\{ Z(\theta)I(x,y) \big\},
\qquad (x,y)\in \mathrm{reg}(D).
\]
}
\end{definition}

Since $Z(\theta)I = -\sin\theta\,\partial_x I + \cos\theta\,\partial_y I$ the above maximization selects the direction for which $Z(\theta)$ aligns with $\nabla I$. Hence $\Theta$ assigns to each point the orientation
orthogonal to the image level set $I(x,y)=$ constant, passing through it. We may think of such image level set as the contour of an object depicted by the image represented by $I$.

To describe border completion, we introduce the horizontal distribution on $\mathcal{E}$:
\[
\mathcal{D}=\Span\{X,Y\},
\qquad 
X=\partial_\theta,
\quad 
Y=\cos\theta\,\partial_x+\sin\theta\,\partial_y .
\]
The word "horizontal" refers to the fact that this distribution is orthogonal to the vector field $Z=[X,Y]$ with respect to the euclidean metric. The fact that the bracket $[X,Y]=Z$ shows that $\mathcal{D}$ is indeed bracket generating: this guarantees the solution of the geodesic existence question (Chow's Thm, see \cite{Montgomery2006ATO}). 

Notice that  $\mathcal{E} \cong \mathrm{SE}(2)$ the special euclidean group, semidirect product of translations
$\R^2$ and rotations $S^1$. Notice furtherly that $\mathcal{E} \longrightarrow \mathcal{E}/S^1\cong \R^2$
is a $S^1$ principal bundle. Endowing $\mathcal{D}$ with the natural invariant metric makes $(\mathcal{E},\mathcal{D})$ a sub-Riemannian manifold, invariant under the action of $\mathrm{SE}(2)\cong \mathcal{E}$ the special euclidean group.

This fact makes the solution to the sub Riemannian geodesic problem unique.

\medskip
Horizontal curves, i.e. curves
tangent to the distribution $\cD$ are
$\Sigma(t)=(x(t),y(t),\theta(t))$ satisfying
\[
\dot\Sigma = u_1 X + u_2 Y,
\]
They represent admissible transitions of position and orientation.
Geodesics minimize their sub-Riemannian length, or equivalently,
in the Hamiltonian formalism, the Hamiltonian
\[
H=\frac12(P_X^2+P_Y^2),
\]
where
\[
P_X=\cos\theta\,p_x+\sin\theta\,p_y,
\qquad
P_Y=p_\theta .
\]
In local coordinates the Hamiltonian reads
\[
H=\frac12\big[(\cos\theta\,p_x+\sin\theta\,p_y)^2 + p_\theta^2\big].
\]

We now compute the geodesics in the Hamiltonian formulation:
\[
\begin{cases}
\dot{x}=\cos\theta\,p_1,\\
\dot{y}=\sin\theta\,p_1,\\
\dot{\theta}=p_2,
\end{cases}
\qquad
\begin{cases}
\dot{p}_1=p_2p_3,\\
\dot{p}_2=-p_1p_3,\\
\dot{p}_3=-p_1p_2,
\end{cases}
\]
where
\[
p_1=P_X,\quad
p_2=P_Y,\quad
p_3=-\sin\theta\,p_x+\cos\theta\,p_y .
\]

Since $H$ is conserved, let $E=2H=p_1^2+p_2^2$.  
Introducing a phase variable $\gamma(t)$ such that
\[
p_1=\sqrt{E}\sin\frac{\gamma}{2},
\qquad
p_2=\sqrt{E}\cos\frac{\gamma}{2},
\qquad
p_3=\tfrac12\dot\gamma,
\]
the projected dynamics become
\[
\begin{cases}
\dot{x}=\sqrt{E}\sin\frac{\gamma}{2}\cos\theta,\\
\dot{y}=\sqrt{E}\sin\frac{\gamma}{2}\sin\theta,\\
\dot{\theta}=\sqrt{E}\cos\frac{\gamma}{2}.
\end{cases}
\]

\begin{figure}
    \centering
    \includegraphics[width=.8\textwidth]{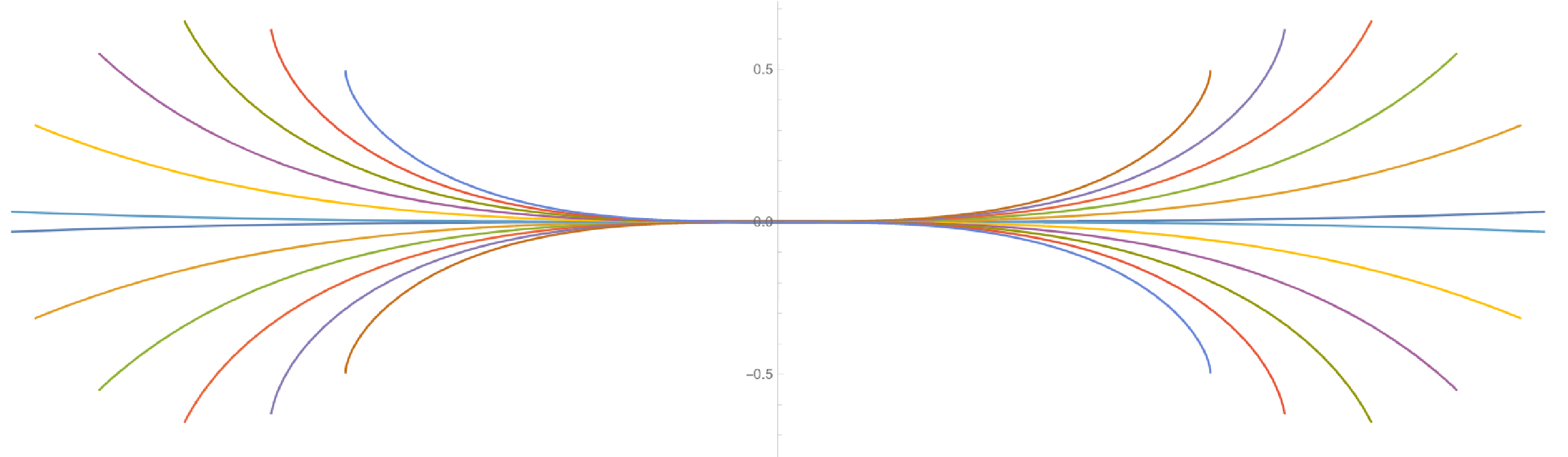}
    \caption{Numerical
      solutions of the geodesic equations for some values of $\gamma$}
    \label{fig::isoenergetic}
\end{figure}

Numerical integration provides the geodesics whose projection in the $x,y$ plane is depicted in Figure \ref{fig::isoenergetic}.
They agree with the celebrated Hayes field as in Figure \ref{Hayes}.

In the next section we shall exploit the modelling of orientation via the vector field $Z$ to enhance border completion capabilities of neural networks.

\section{Experiments}

We create a new CNN model in which custom filters are added by mimicking the orientation map \eqref{ormap} in our mathematical modelling. 

\begin{figure}[h!]
  \centering
  \includegraphics[width=0.9\textwidth]{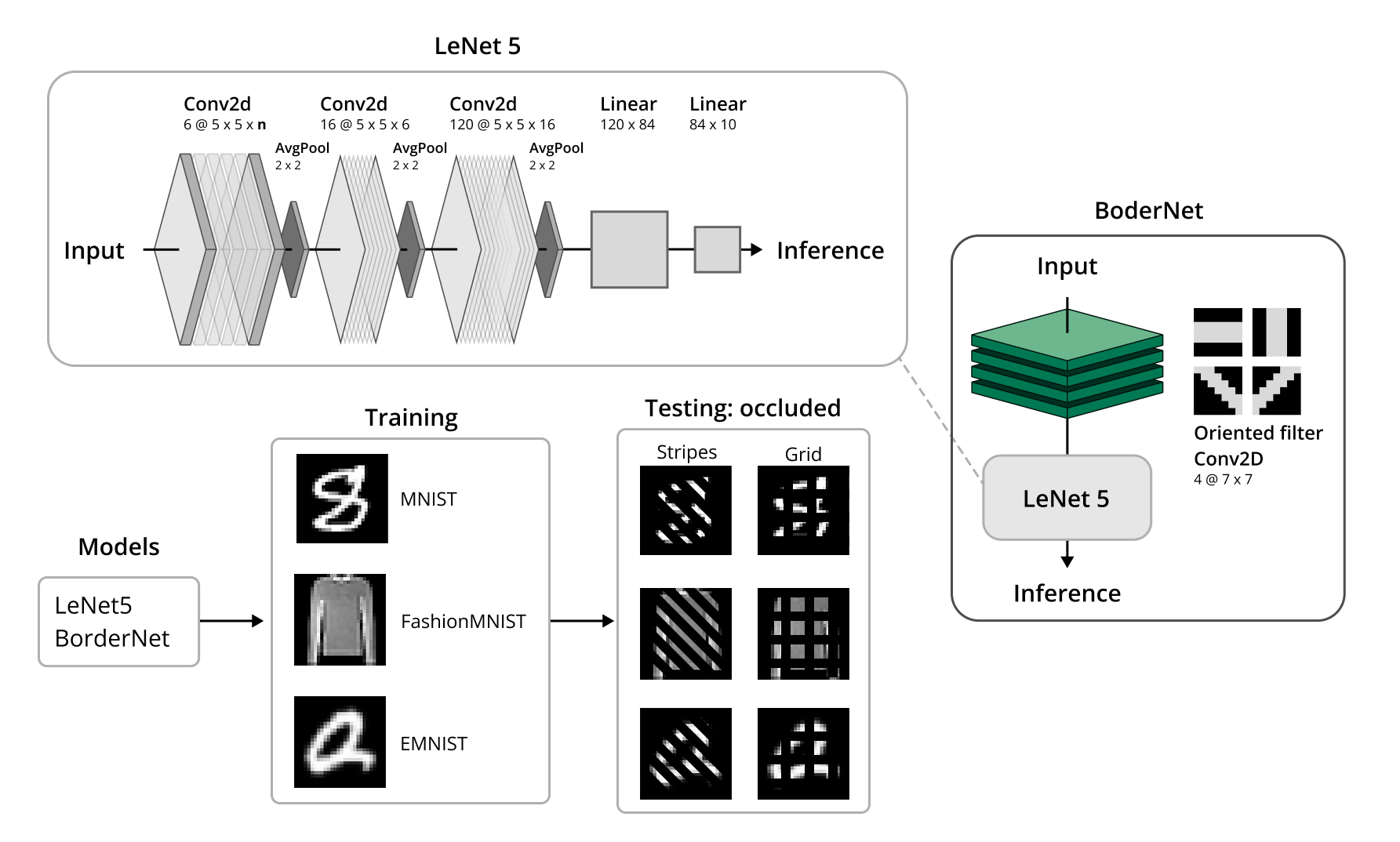}
  \caption{Implemented pipeline with Vanilla LeNet5 and BorderNet trained on unoccluded images of the three different datasets (MNIST/FashionMNIST/EMNIST) and tested on two types of occluded images (stripes/grid). }
  \label{fig:pipeline}
\end{figure}

\begin{figure}
\begin{center}
  \includegraphics[width=.95\textwidth]{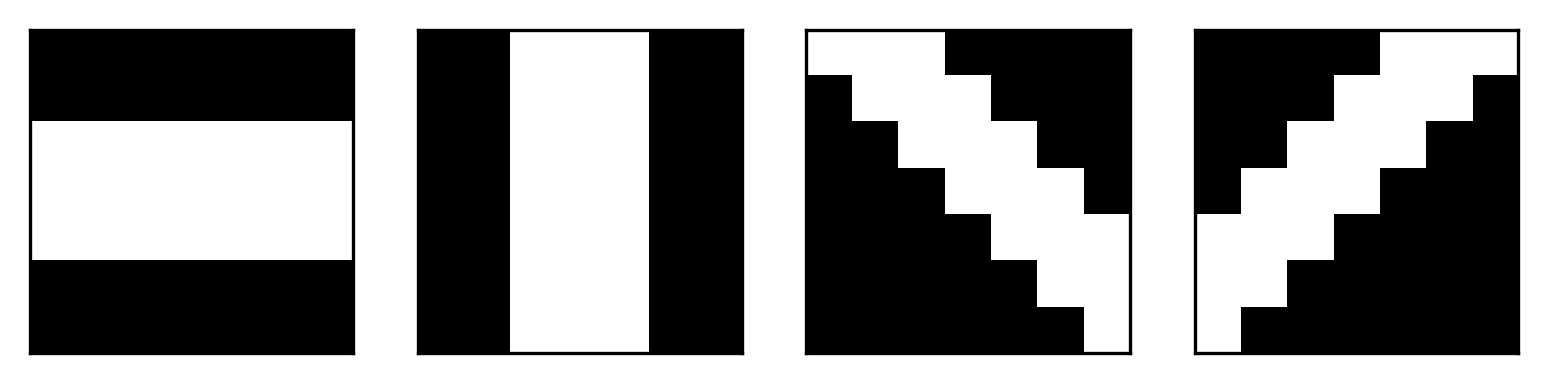}
\end{center}
  \caption{BorderNet oriented filters.} 
  \label{fig:filters}
\end{figure}

\medskip
{\bf CNN models.}
We considered two CNN models in our study: the LeNet5 (Vanilla LeNet5) and a modified LeNet5, called BorderNet, in which four custom filters were added at the beginning as a sequence of convolutions: see Figure \ref{fig:pipeline} for a scheme describing the structure and differences of LeNet5 and BorderNet. The filters of BorderNet were created based on the analogy with the visual cortex, each filter featuring one direction - horizontal, vertical, and both diagonals (see Figure~\ref{fig:filters}). Filters were defined with a size of $7\times 7$ pixels and a stripe width
of 3 pixels, where the pixels belonging to the oriented stripe were set to the value of $1$, while the remaining background was set to $0$. In this fashion, a convolution with the filter would mimics the action of the vector field $Z$ as described above.

\medskip
{\bf Datasets.}
We examine the performance of the two CNN models on three datasets: MNIST, Fashion MNIST and Extended MNIST digits (EMNIST). For each dataset, both models were trained on unoccluded images using the ADAM optimizer, with a learning rate of 0.001, 10 epochs, and a batch size of 64. The testing was carried out using the occluded  images. To reduce variability inherent to our models, we performed 100 cycles for each experiment, using a fixed random seed (seed=42) to ensure reproducibility. Individual model accuracies were calculated as the mean over the 100 cycles for each combination of occlusion stripe width and spacing. For model comparison, we computed an accuracy improvement of BorderNet with respect to LeNet5 which does not follow a normal distribution. We therefore chose to express it as the median improvement, associated with 95\% confidence intervals bootstrapped with 100000 bootstrap samples.

\medskip

{\bf Occlusions in Test Datasets.}
To test robustness of our modified LeNet5, i.e. BorderNet, we created test datasets  consisting of two types of occluded images as in Fig. \ref{fig:occlusions}. For stripe occlusions we take a masking composed of diagonal straight stripes of width $w$ and inter-stripe spacing $s$, whereas for grid occlusions instead of diagonal stripes
we considered square grids composed of horizontal and vertical stripes. To conduct a thorough benchmarking, we test all the models on all the occlusions obtained from the pairwise combination of $s, w \in [1,10]$ for each dataset separately (see Figure \ref{fig:pipeline} for a scheme of our pipeline). 

{
\begin{figure}[h!]
 \centering
     \subfloat[\centering Stripe Occlusions]{{\includegraphics[width=0.8\textwidth]{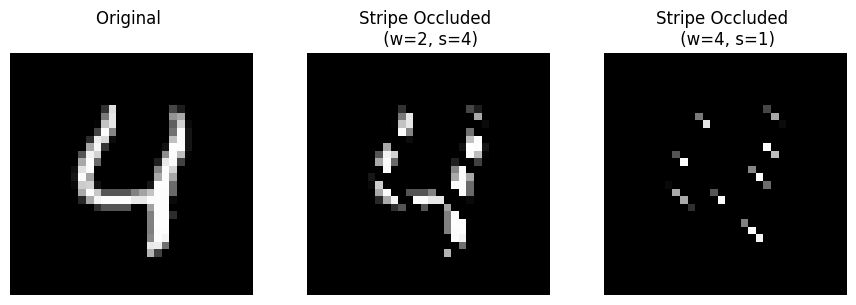} }}%
    \qquad
     \\ [1.5ex] 
     \subfloat[\centering Grid Occlusions]{{\includegraphics[width=0.8\textwidth]{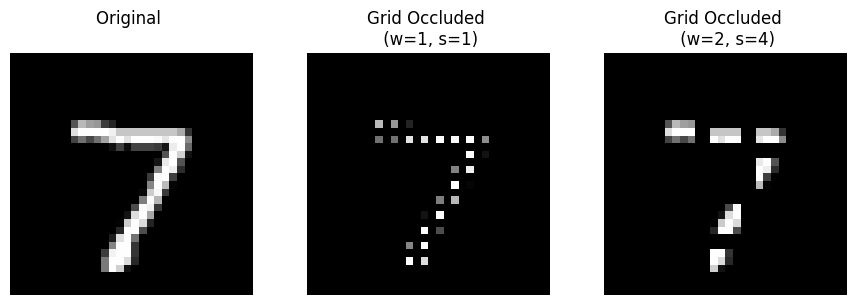} }}%
     \caption{Sample of images occluded with stripes (top row) and grids (bottom row) used in the testing phase occluded.} 
    \label{fig:occlusions}%
 \end{figure}

\section{Results}\label{sec-res}

LeNet5 and BorderNet accuracies are reported in this section. Table 1 lists the mean accuracies of our models, separately for the MNIST, FashionMNIST and EMNIST images occluded with stripes. Table 2 reports the same outcomes but for grid occlusions. Boostrapped median improvement comparison of BorderNet over LeNet5 are also reported. 

\medskip

\begin{table}[h!]
\caption{Mean Accuracies of Vanilla LeNet5 and BorderNet for some examples of \textbf{stripe} occlusions for MNIST/FashionMNIST/EMNIST for different combinations of stripe width and stripe spacing \textit{(w,s)}. Improvement (Imp.*) of BorderNet over Vanilla LeNet5 are the bootstrapped medians in percentage.}
\vspace{2mm}
\label{tab:stripe_acc}
\scriptsize
\begin{tabular}{l|ccc|ccc|ccc|}
\cline{2-10}
 &
  \multicolumn{3}{c|}{\textbf{MNIST}} &
  \multicolumn{3}{c|}{\textbf{FashionMNIST}} &
  \multicolumn{3}{c|}{\textbf{EMNIST}} \\ \hline
\multicolumn{1}{|l|}{\textit{(w,s)}} &
  \multicolumn{1}{l|}{\textbf{LeNet5}} &
  \multicolumn{1}{l|}{\textbf{BorderNet}} &
  \multicolumn{1}{l|}{\textbf{Imp.*}} &
  \multicolumn{1}{l|}{\textbf{LeNet5}} &
  \multicolumn{1}{l|}{\textbf{BorderNet}} &
  \multicolumn{1}{l|}{\textbf{Imp.*}} &
  \multicolumn{1}{l|}{\textbf{LeNet5}} &
  \multicolumn{1}{l|}{\textbf{BorderNet}} &
  \multicolumn{1}{l|}{\textbf{Imp.*}} \\ \hline
\multicolumn{1}{|l|}{(1,1)} &
  \multicolumn{1}{c|}{0.974} &
  \multicolumn{1}{c|}{0.974} &
  99.90\% &
  \multicolumn{1}{c|}{0.848} &
  \multicolumn{1}{c|}{0.853} &
  100.60\% &
  \multicolumn{1}{c|}{0.989} &
  \multicolumn{1}{c|}{0.985} &
  99.70\% \\ \hline
\multicolumn{1}{|l|}{(1,10)} &
  \multicolumn{1}{c|}{0.982} &
  \multicolumn{1}{c|}{0.979} &
  99.70\% &
  \multicolumn{1}{c|}{0.854} &
  \multicolumn{1}{c|}{0.867} &
  101.50\% &
  \multicolumn{1}{c|}{0.991} &
  \multicolumn{1}{c|}{0.988} &
  99.70\% \\ \hline
\multicolumn{1}{|l|}{(2,4)} &
  \multicolumn{1}{c|}{0.939} &
  \multicolumn{1}{c|}{0.960} &
  101.80\% &
  \multicolumn{1}{c|}{0.822} &
  \multicolumn{1}{c|}{0.774} &
  95.20\% &
  \multicolumn{1}{c|}{0.962} &
  \multicolumn{1}{c|}{0.976} &
  101.30\% \\ \hline
\multicolumn{1}{|l|}{(3,3)} &
  \multicolumn{1}{c|}{0.788} &
  \multicolumn{1}{c|}{0.839} &
  105.90\% &
  \multicolumn{1}{c|}{0.689} &
  \multicolumn{1}{c|}{0.683} &
  100.70\% &
  \multicolumn{1}{c|}{0.843} &
  \multicolumn{1}{c|}{0.891} &
  105.20\% \\ \hline
\multicolumn{1}{|l|}{(4,1)} &
  \multicolumn{1}{c|}{0.573} &
  \multicolumn{1}{c|}{0.630} &
  108.90\% &
  \multicolumn{1}{c|}{0.514} &
  \multicolumn{1}{c|}{0.523} &
  101.20\% &
  \multicolumn{1}{c|}{0.673} &
  \multicolumn{1}{c|}{0.697} &
  103.70\% \\ \hline
\multicolumn{1}{|l|}{(5,3)} &
  \multicolumn{1}{c|}{0.419} &
  \multicolumn{1}{c|}{0.426} &
  101.40\% &
  \multicolumn{1}{c|}{0.362} &
  \multicolumn{1}{c|}{0.375} &
  103.60\% &
  \multicolumn{1}{c|}{0.412} &
  \multicolumn{1}{c|}{0.603} &
  147.10\% \\ \hline
\multicolumn{1}{|l|}{(6,4)} &
  \multicolumn{1}{c|}{0.354} &
  \multicolumn{1}{c|}{0.313} &
  87.60\% &
  \multicolumn{1}{c|}{0.354} &
  \multicolumn{1}{c|}{0.385} &
  112.50\% &
  \multicolumn{1}{c|}{0.322} &
  \multicolumn{1}{c|}{0.477} &
  147.60\% \\ \hline
\multicolumn{1}{|l|}{(7,9)} &
  \multicolumn{1}{c|}{0.362} &
  \multicolumn{1}{c|}{0.389} &
  108.30\% &
  \multicolumn{1}{c|}{0.322} &
  \multicolumn{1}{c|}{0.268} &
  83.90\% &
  \multicolumn{1}{c|}{0.446} &
  \multicolumn{1}{c|}{0.466} &
  105.80\% \\ \hline
\multicolumn{1}{|l|}{(8,6)} &
  \multicolumn{1}{c|}{0.291} &
  \multicolumn{1}{c|}{0.321} &
  109.70\% &
  \multicolumn{1}{c|}{0.253} &
  \multicolumn{1}{c|}{0.230} &
  91.40\% &
  \multicolumn{1}{c|}{0.338} &
  \multicolumn{1}{c|}{0.366} &
  109.90\% \\ \hline
\multicolumn{1}{|l|}{(9,3)} &
  \multicolumn{1}{c|}{0.229} &
  \multicolumn{1}{c|}{0.215} &
  91.00\% &
  \multicolumn{1}{c|}{0.223} &
  \multicolumn{1}{c|}{0.212} &
  96.60\% &
  \multicolumn{1}{c|}{0.224} &
  \multicolumn{1}{c|}{0.251} &
  112.50\% \\ \hline
\multicolumn{1}{|l|}{(10,1)} &
  \multicolumn{1}{c|}{0.178} &
  \multicolumn{1}{c|}{0.142} &
  76.70\% &
  \multicolumn{1}{c|}{0.156} &
  \multicolumn{1}{c|}{0.206} &
  139.40\% &
  \multicolumn{1}{c|}{0.167} &
  \multicolumn{1}{c|}{0.170} &
  100.70\% \\ \hline
\multicolumn{1}{|l|}{(10,10)} &
  \multicolumn{1}{c|}{0.178} &
  \multicolumn{1}{c|}{0.242} &
  132.90\% &
  \multicolumn{1}{c|}{0.170} &
  \multicolumn{1}{c|}{0.309} &
  186.80\% &
  \multicolumn{1}{c|}{0.282} &
  \multicolumn{1}{c|}{0.326} &
  119.50\% \\ \hline
\end{tabular}
\end{table}

\vspace*{-6mm}

\begin{table}[h!]
\caption{Mean Accuracies of Vanilla LeNet5 and BorderNet for some examples of \textbf{grid} occlusions for MNIST/FashionMNIST/EMNIST for different combinations of stripe width and stripe spacing \textit{(w,s)}. Improvement (Imp.*) of BorderNet over Vanilla LeNet5 are the bootstrapped medians in percentage.}
\vspace{2mm}
\label{tab:grid_acc}
\scriptsize
\begin{tabular}{l|ccl|ccl|ccl|}
\cline{2-10}
                                     & \multicolumn{3}{c|}{\textbf{MNIST}}                                                             & \multicolumn{3}{c|}{\textbf{FashionMNIST}}                                                      & \multicolumn{3}{c|}{\textbf{EMNIST}}                                                            \\ \hline
\multicolumn{1}{|l|}{\textit{(w,s)}} & \multicolumn{1}{l|}{\textbf{LeNet5}} & \multicolumn{1}{l|}{\textbf{BorderNet}} & \textbf{Imp.*} & \multicolumn{1}{l|}{\textbf{LeNet5}} & \multicolumn{1}{l|}{\textbf{BorderNet}} & \textbf{Imp.*} & \multicolumn{1}{l|}{\textbf{LeNet5}} & \multicolumn{1}{l|}{\textbf{BorderNet}} & \textbf{Imp.*} \\ \hline
\multicolumn{1}{|l|}{(1,1)}          & \multicolumn{1}{c|}{0.882}           & \multicolumn{1}{c|}{0.859}              & 95.50\%        & \multicolumn{1}{c|}{0.722}           & \multicolumn{1}{c|}{0.717}              & 99.60\%        & \multicolumn{1}{c|}{0.939}           & \multicolumn{1}{c|}{0.890}              & 94.10\%        \\ \hline
\multicolumn{1}{|l|}{(1,10)}         & \multicolumn{1}{c|}{0.966}           & \multicolumn{1}{c|}{0.973}              & 106.60\%       & \multicolumn{1}{c|}{0.823}           & \multicolumn{1}{c|}{0.852}              & 103.40\%       & \multicolumn{1}{c|}{0.979}           & \multicolumn{1}{c|}{0.984}              & 100.50\%       \\ \hline
\multicolumn{1}{|l|}{(2,4)}          & \multicolumn{1}{c|}{0.674}           & \multicolumn{1}{c|}{0.764}              & 112.70\%       & \multicolumn{1}{c|}{0.549}           & \multicolumn{1}{c|}{0.675}              & 123.30\%       & \multicolumn{1}{c|}{0.647}           & \multicolumn{1}{c|}{0.811}              & 125.80\%       \\ \hline
\multicolumn{1}{|l|}{(3,3)}          & \multicolumn{1}{c|}{0.384}           & \multicolumn{1}{c|}{0.412}              & 107.50\%       & \multicolumn{1}{c|}{0.381}           & \multicolumn{1}{c|}{0.433}              & 113.40\%       & \multicolumn{1}{c|}{0.398}           & \multicolumn{1}{c|}{0.413}              & 102.00\%       \\ \hline
\multicolumn{1}{|l|}{(4,1)}          & \multicolumn{1}{c|}{0.153}           & \multicolumn{1}{c|}{0.234}              & 152.40\%       & \multicolumn{1}{c|}{0.174}           & \multicolumn{1}{c|}{0.304}              & 177.50\%       & \multicolumn{1}{c|}{0.143}           & \multicolumn{1}{c|}{0.168}              & 120.20\%       \\ \hline
\multicolumn{1}{|l|}{(5,3)}          & \multicolumn{1}{c|}{0.211}           & \multicolumn{1}{c|}{0.205}              & 95.10\%        & \multicolumn{1}{c|}{0.300}           & \multicolumn{1}{c|}{0.258}              & 87.80\%        & \multicolumn{1}{c|}{0.274}           & \multicolumn{1}{c|}{0.292}              & 104.10\%       \\ \hline
\multicolumn{1}{|l|}{(6,4)}          & \multicolumn{1}{c|}{0.250}           & \multicolumn{1}{c|}{0.198}              & 77.20\%        & \multicolumn{1}{c|}{0.163}           & \multicolumn{1}{c|}{0.243}              & 155.90\%       & \multicolumn{1}{c|}{0.207}           & \multicolumn{1}{c|}{0.219}              & 105.10\%       \\ \hline
\multicolumn{1}{|l|}{(7,9)}          & \multicolumn{1}{c|}{0.232}           & \multicolumn{1}{c|}{0.285}              & 122.20\%       & \multicolumn{1}{c|}{0.317}           & \multicolumn{1}{c|}{0.304}              & 96.60\%        & \multicolumn{1}{c|}{0.319}           & \multicolumn{1}{c|}{0.351}              & 108.20\%       \\ \hline
\multicolumn{1}{|l|}{(8,6)}          & \multicolumn{1}{c|}{0.182}           & \multicolumn{1}{c|}{0.209}              & 115.00\%       & \multicolumn{1}{c|}{0.235}           & \multicolumn{1}{c|}{0.240}              & 102.60\%       & \multicolumn{1}{c|}{0.259}           & \multicolumn{1}{c|}{0.258}              & 99.70\%        \\ \hline
\multicolumn{1}{|l|}{(9,3)}          & \multicolumn{1}{c|}{0.158}           & \multicolumn{1}{c|}{0.145}              & 86.60\%        & \multicolumn{1}{c|}{0.195}           & \multicolumn{1}{c|}{0.177}              & 88.90\%        & \multicolumn{1}{c|}{0.168}           & \multicolumn{1}{c|}{0.158}              & 95.60\%        \\ \hline
\multicolumn{1}{|l|}{(10,1)}         & \multicolumn{1}{c|}{0.117}           & \multicolumn{1}{c|}{0.106}              & 74.00\%        & \multicolumn{1}{c|}{0.125}           & \multicolumn{1}{c|}{0.166}              & 168.00\%       & \multicolumn{1}{c|}{0.108}           & \multicolumn{1}{c|}{0.110}              & 100.00\%       \\ \hline
\multicolumn{1}{|l|}{(10,10)}        & \multicolumn{1}{c|}{0.372}           & \multicolumn{1}{c|}{0.366}              & 97.20\%        & \multicolumn{1}{c|}{0.241}           & \multicolumn{1}{c|}{0.183}              & 72.10\%        & \multicolumn{1}{c|}{0.279}           & \multicolumn{1}{c|}{0.225}              & 81.40\%        \\ \hline
\end{tabular}
\end{table}
}

\newpage

\begin{figure}[h!]
 \centering
     \subfloat[\centering MNIST Stripe Occlusions]{{\includegraphics[width=0.9\textwidth]{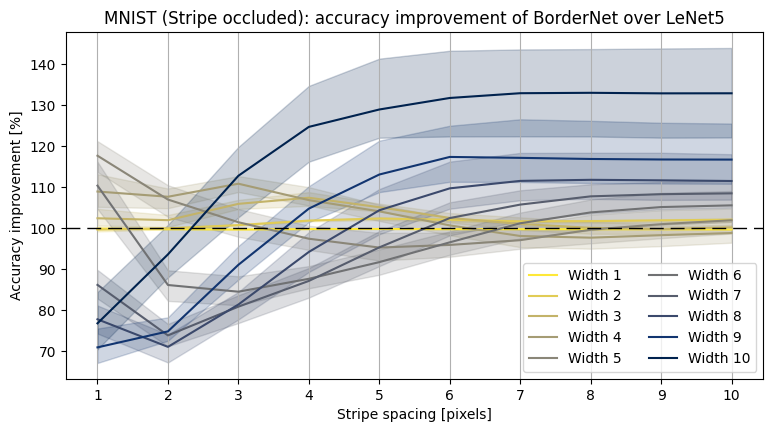} }}%
     \qquad
     \\ [1.5ex] 
     \subfloat[\centering MNIST Grid Occlusions]{{\includegraphics[width=0.9\textwidth]{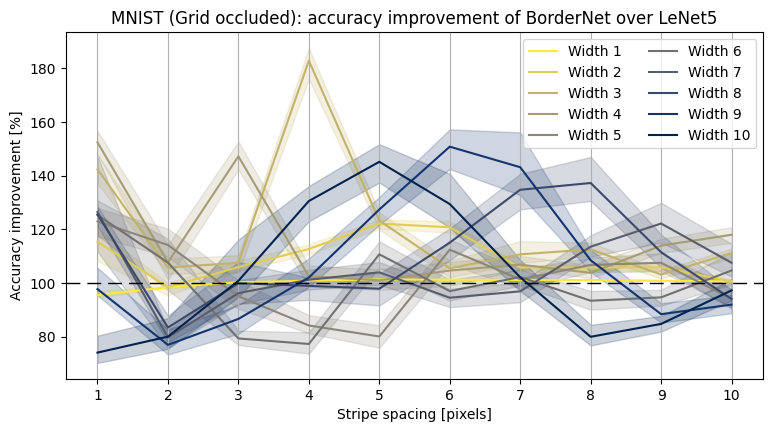} }}%
    \caption{Accuracy improvement of BorderNet with respect to Vanilla LeNet5 for occluded MNIST images with stripes (a) and grid (b).} 
     \label{fig:MNIST_imp}%
 \end{figure}

\begin{figure}[h!]
 \centering
     \subfloat[\centering FashionMNIST Stripe Occlusions]{{\includegraphics[width=0.9\textwidth]{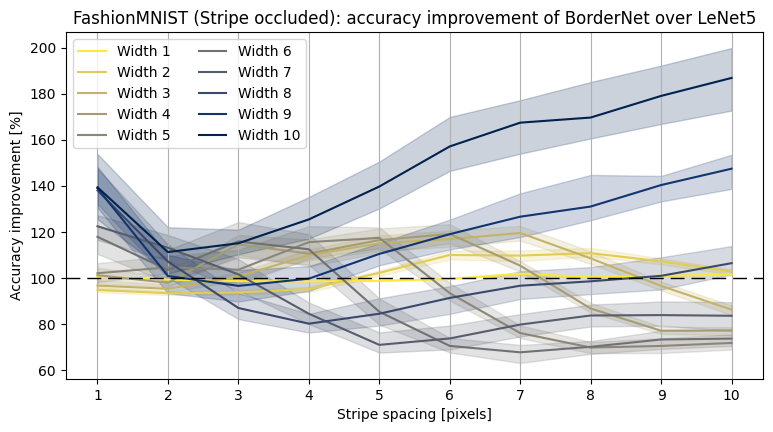} }}
     \qquad
     \\ [1.5ex] 
     \subfloat[\centering FashionMNIST Grid Occlusions]{{\includegraphics[width=0.9\textwidth]{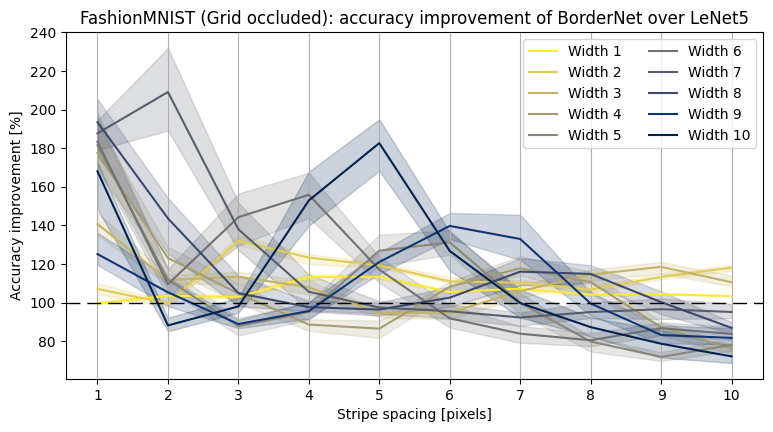} }}%
    \caption{Accuracy improvement of BorderNet with respect to Vanilla LeNet5 for occluded FashionMNIST images with stripes (a) and grid (b).} 
     \label{fig:FashionMNIST_imp}%
 \end{figure}

We see a consistent improvement  when comparing BorderNet with LeNet5 
except in the case of severe occlusions, where no comparison can be
significant. We also depict graphically the improvement comparison results for all widths and stripe spacing for the MNIST (Figure \ref{fig:MNIST_imp}), FashionMNIST (Figure \ref{fig:FashionMNIST_imp}) and EMNIST datasets (Figure \ref{fig:EMNIST_imp}). Single model's performance graphs for each type of occlusions can be found in the Supplementary Material.

 \begin{figure}[h!]
 \centering
     \subfloat[\centering EMNIST Stripe Occlusions]{{\includegraphics[width=0.9\textwidth]{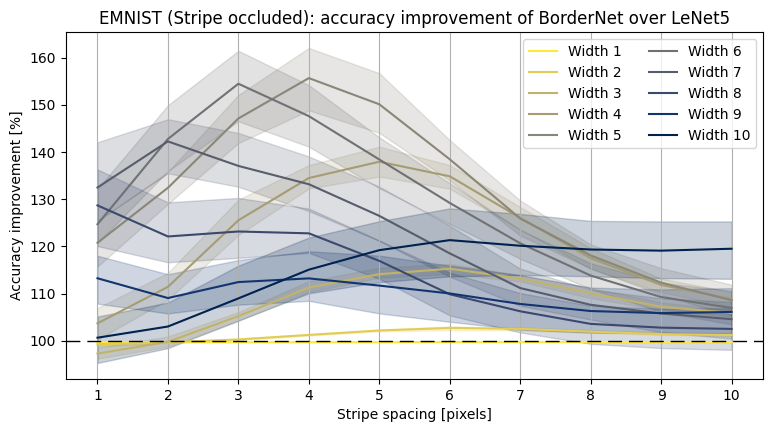} }}%
     \qquad
     \\ [1.5ex]
     \subfloat[\centering EMNIST Grid Occlusions]{{\includegraphics[width=0.9\textwidth]{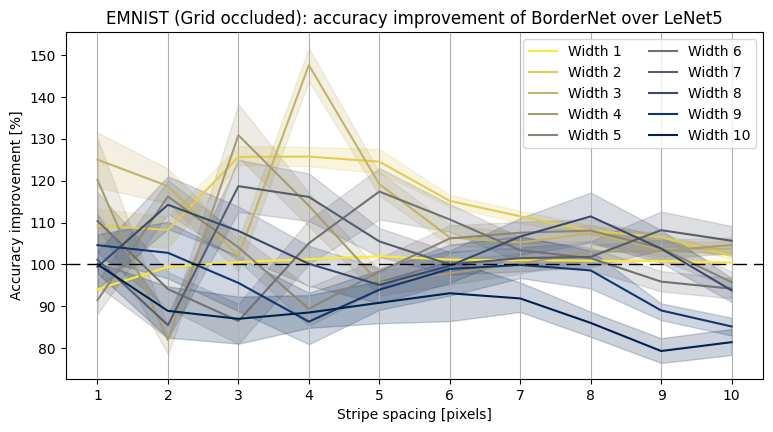} }}%
    \caption{Accuracy improvement of BorderNet with respect to Vanilla LeNet5 for occluded EMNIST images with stripes (a) and grid (b).} 
     \label{fig:EMNIST_imp}%
 \end{figure}

\clearpage

\section{Conclusions}

We show how BorderNet, a CNN model with bioinspired filters, performs consistently better than LeNet5 on occluded test images from three different datasets: MNIST, FashionMNIST and EMNIST and with respect to two diverse sets of occlusions, stripes and grids. This confirms the proof of concept result expressed in \cite{Coutinho2025EnhancingCR} and suggest future directions in bioinspired CNNs.

\begin{credits}
\subsubsection{\ackname}

We thank Prof. R. Duits for helpful comments and observations regarding our work.

The research of the first author was supported by a research scholarship funded by the European Union-Next Generation EU, Missione 4 Component 1 CUP J33C24001310009.
This research was funded by CaLISTA CA 21109, CaLIGOLA MSCA-2021-SE-01-101086123, MSCA-DN CaLiForNIA—101119552, PNRR MNESYS, the PNRR National Center for HPC, Big Data, and Quantum Computing, SimQuSec; INFN Sezione Bologna, Gast initiative and GNSAGA Indam.

\subsubsection{\discintname}
The authors have no competing interests.
\end{credits}

\bibliographystyle{splncs04}
\bibliography{biblio}

\clearpage
\appendix
\section*{Supplementary Material}\label{sup}
\addcontentsline{toc}{section}{Supplementary Material}

For each dataset (MNIST/FashionMNIST/EMNIST) and occlusion type (grid/stripe) plots of mean accuracies and standard deviation for Vanilla LeNet5 and BorderNet are presented. 

\begin{figure}[h]
  \centering
  \begin{minipage}{0.5\textwidth}
    \centering
    \includegraphics[width=\linewidth]{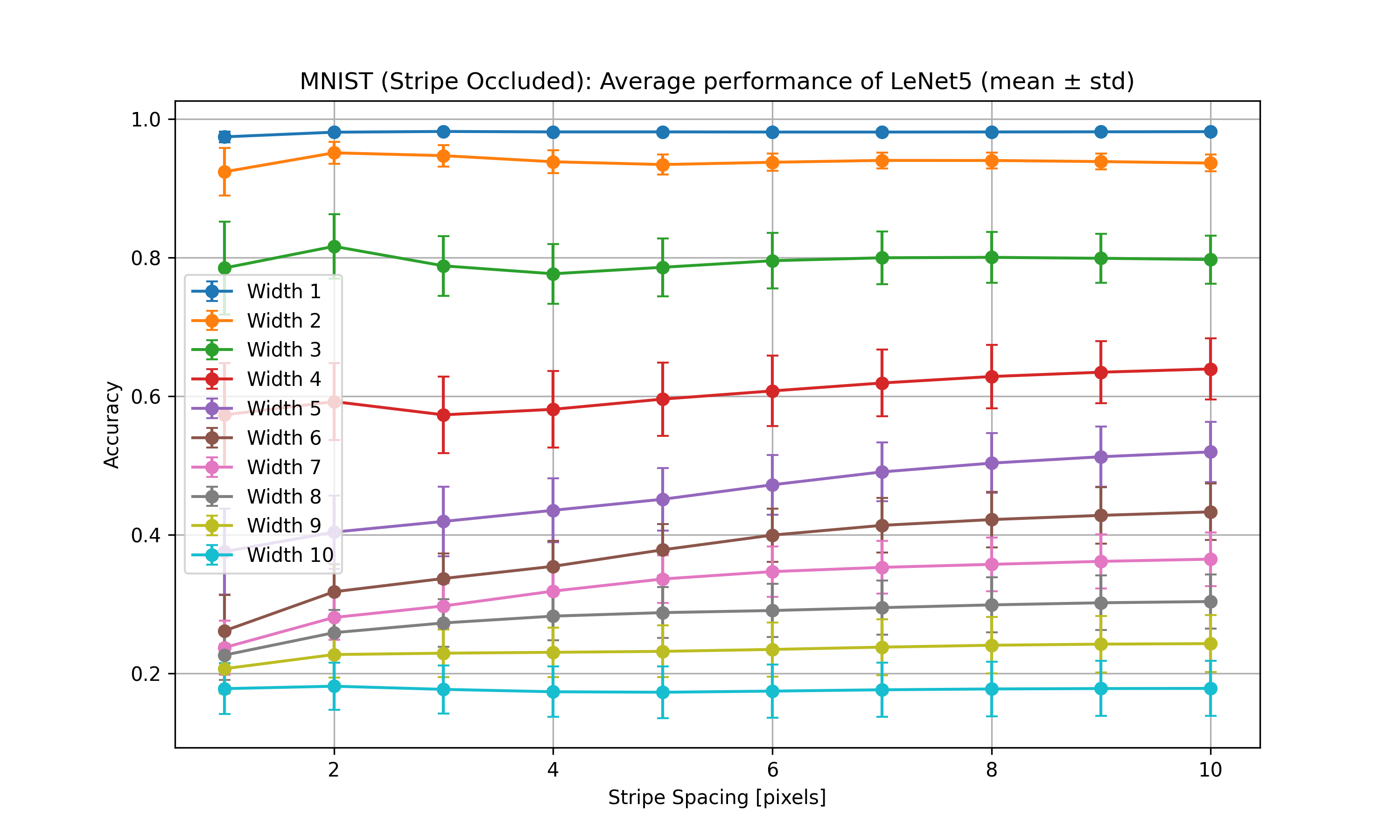}
    \caption*{(a) Vanilla LeNet5}  
  \end{minipage}%
  \begin{minipage}{0.5\textwidth}
    \centering
    \includegraphics[width=\linewidth]{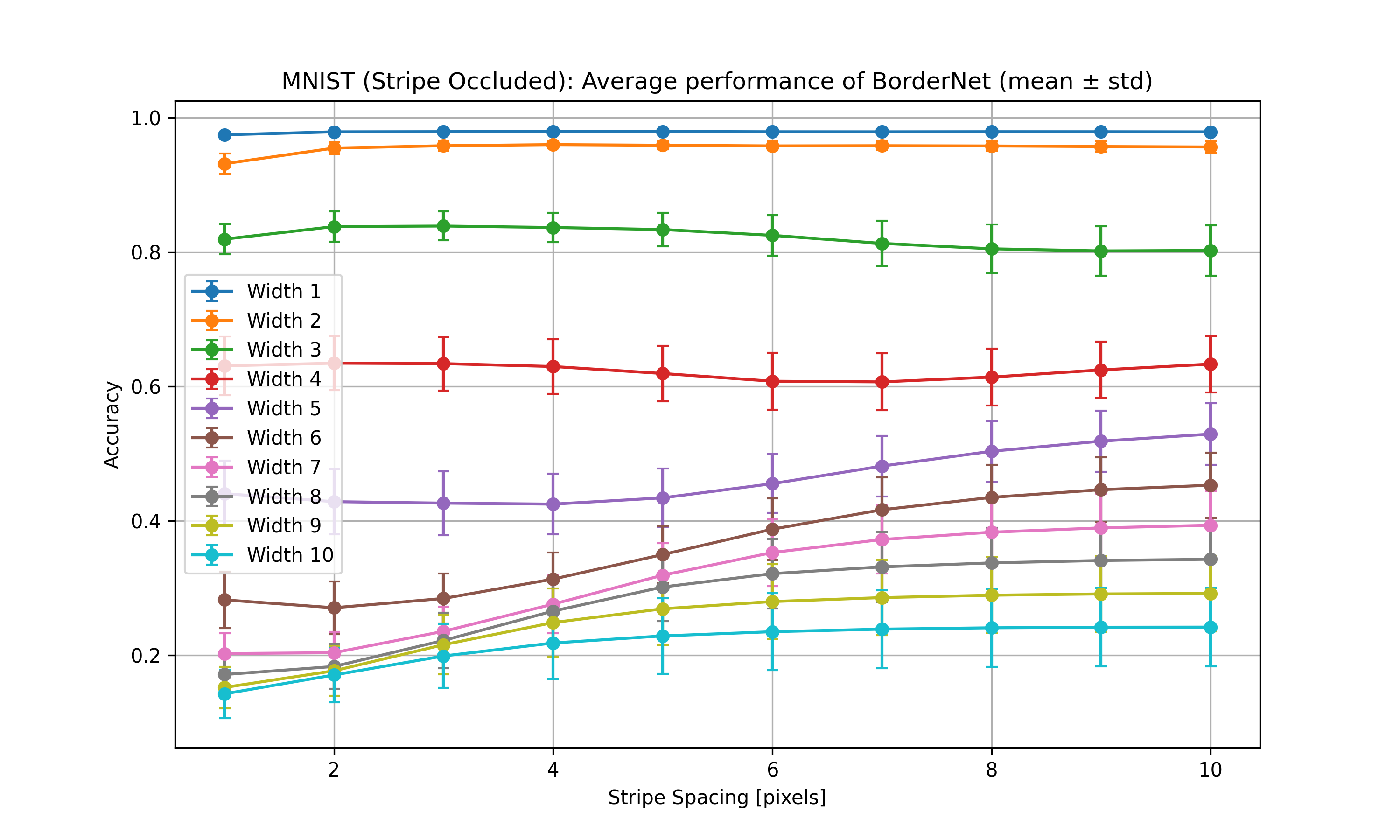}
    \caption*{(b) BorderNet} 
  \end{minipage}
  \caption{Accuracies with MNIST occlusion stripe spacing and width for the Vanilla LeNet5 and BorderNet.}  
  \label{fig:MNIST_stripe_perf}
\end{figure}

\vspace*{-10mm}

\begin{figure}[h]
  \centering
  \begin{minipage}{0.5\textwidth}
    \centering
    \includegraphics[width=\linewidth]{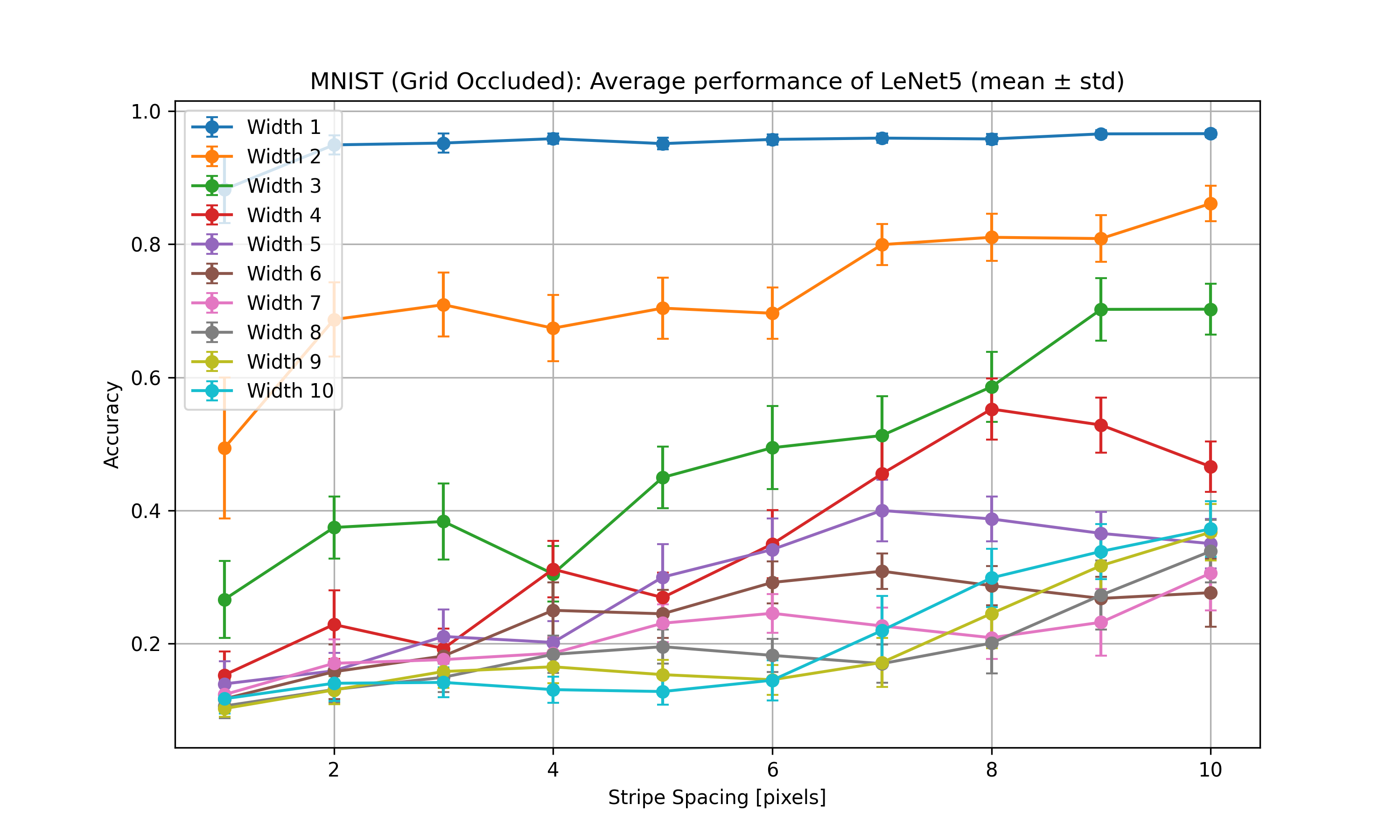}
    \caption*{(a) Vanilla LeNet5}  
  \end{minipage}%
  \begin{minipage}{0.5\textwidth}
    \centering
    \includegraphics[width=\linewidth]{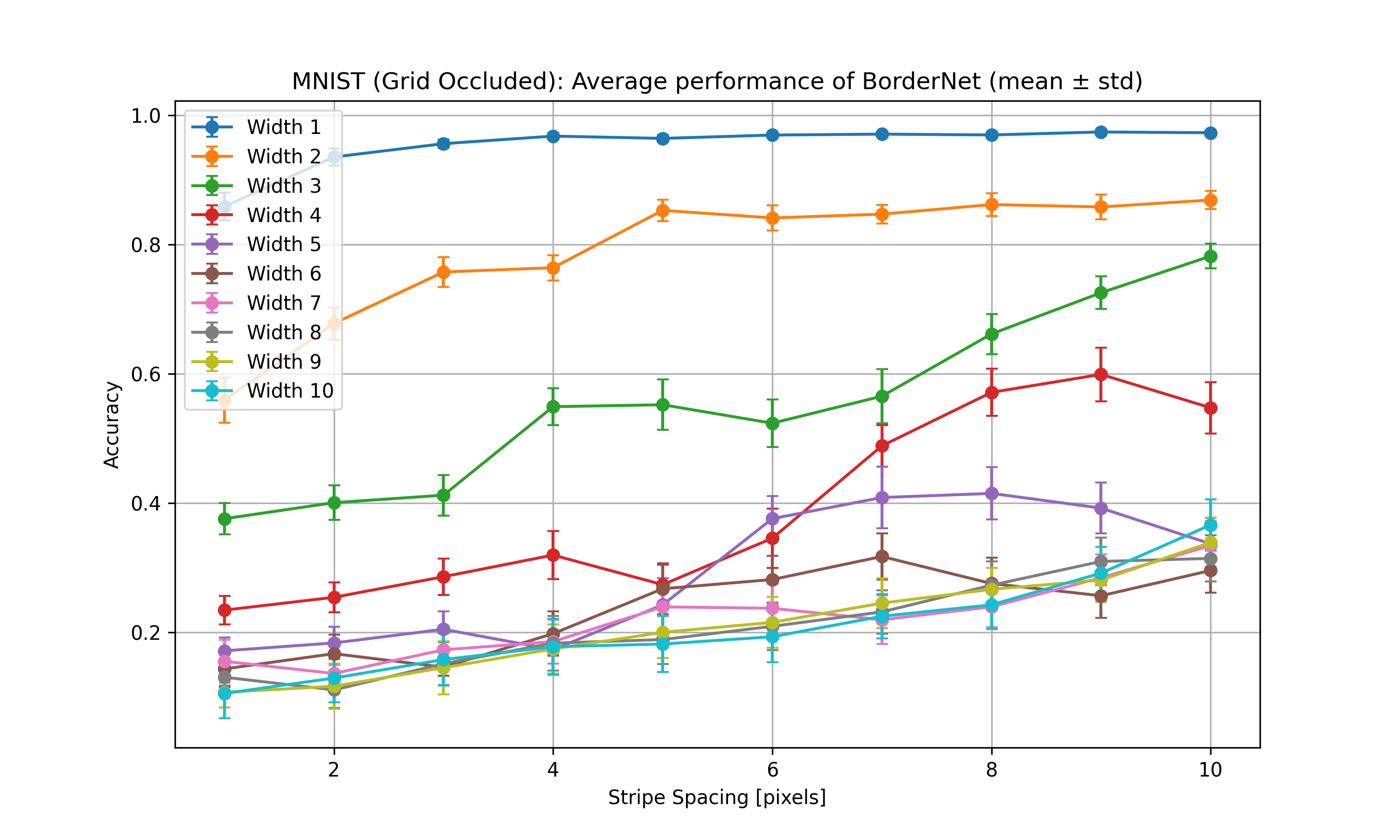}
    \caption*{(b) BorderNet} 
  \end{minipage}
  \caption{Accuracies with MNIST occlusion grid spacing and width for the Vanilla LeNet5 and BorderNet.}  
  \label{fig:MNIST_grid_perf}
\end{figure}

\vspace*{-10mm}

\begin{figure}[h]
  \centering
  \begin{minipage}{0.5\textwidth}
    \centering
    \includegraphics[width=\linewidth]{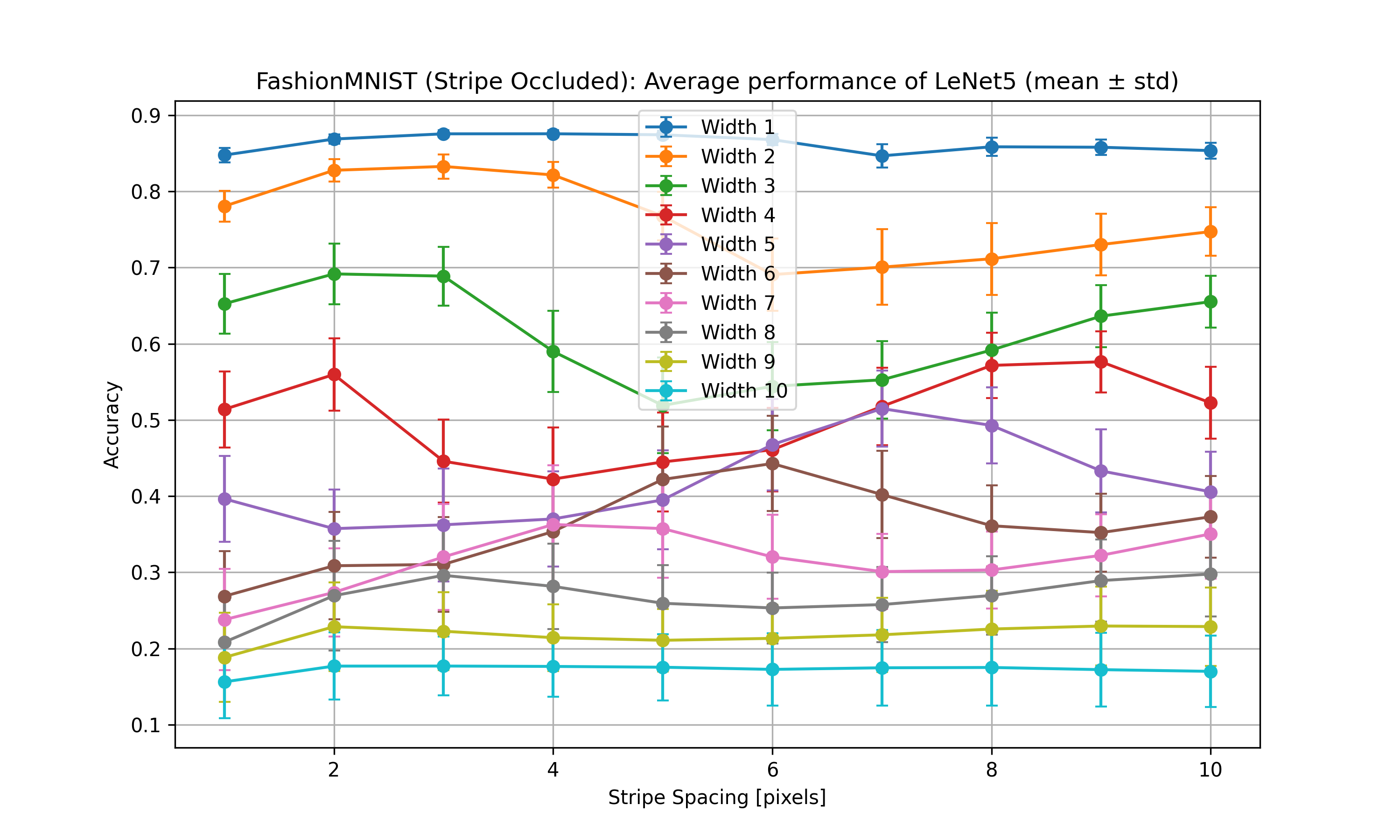}
    \caption*{(a) Vanilla LeNet5}  
  \end{minipage}%
  \begin{minipage}{0.5\textwidth}
    \centering
    \includegraphics[width=\linewidth]{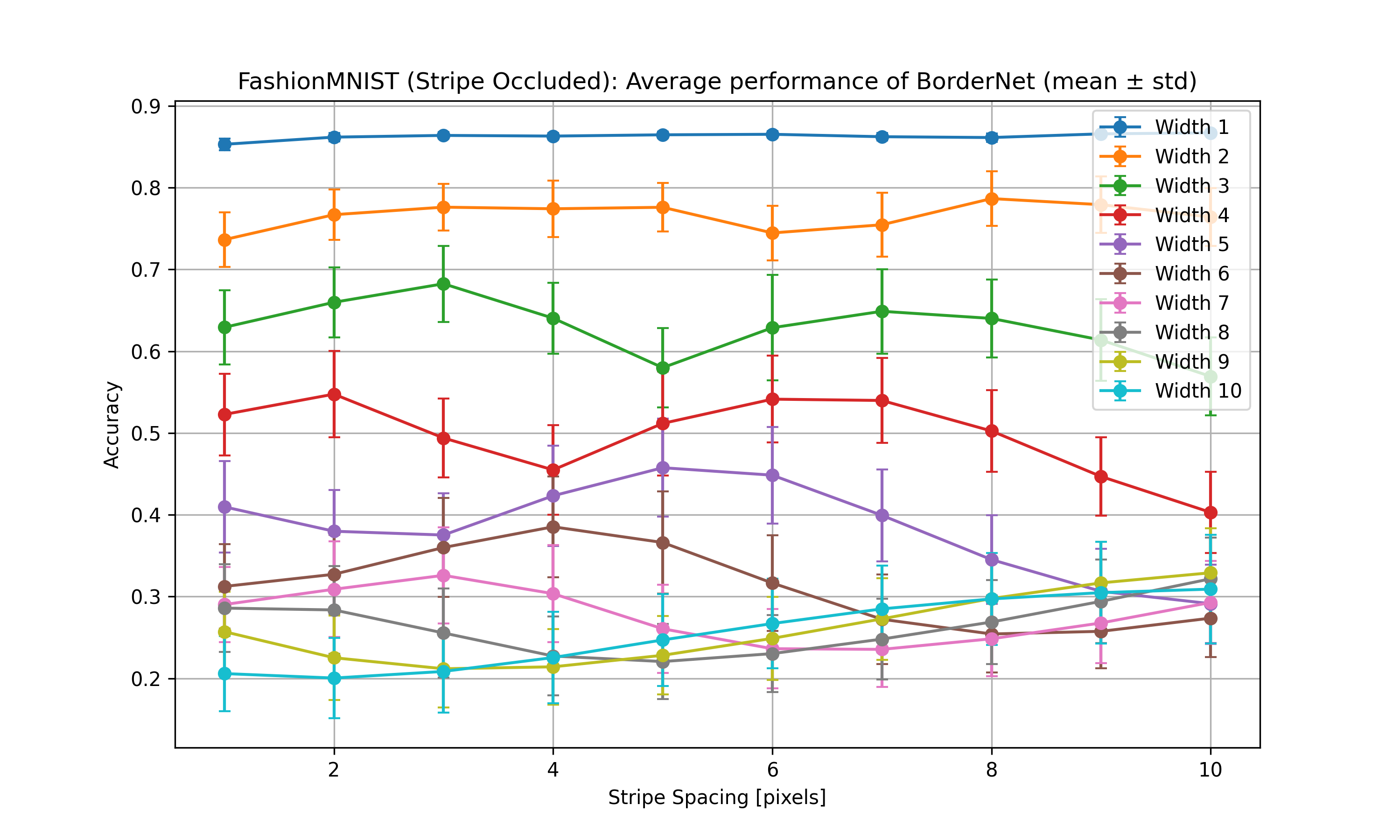}
    \caption*{(b) BorderNet} 
  \end{minipage}
  \caption{Accuracies with FashionMNIST occlusion stripe spacing and width for the Vanilla LeNet5 and BorderNet.}  
  \label{fig:FashionMNIST_stripe_perf}
\end{figure}

\vspace*{-10mm}

\begin{figure}[h]
  \centering
  \begin{minipage}{0.5\textwidth}
    \centering
    \includegraphics[width=\linewidth]{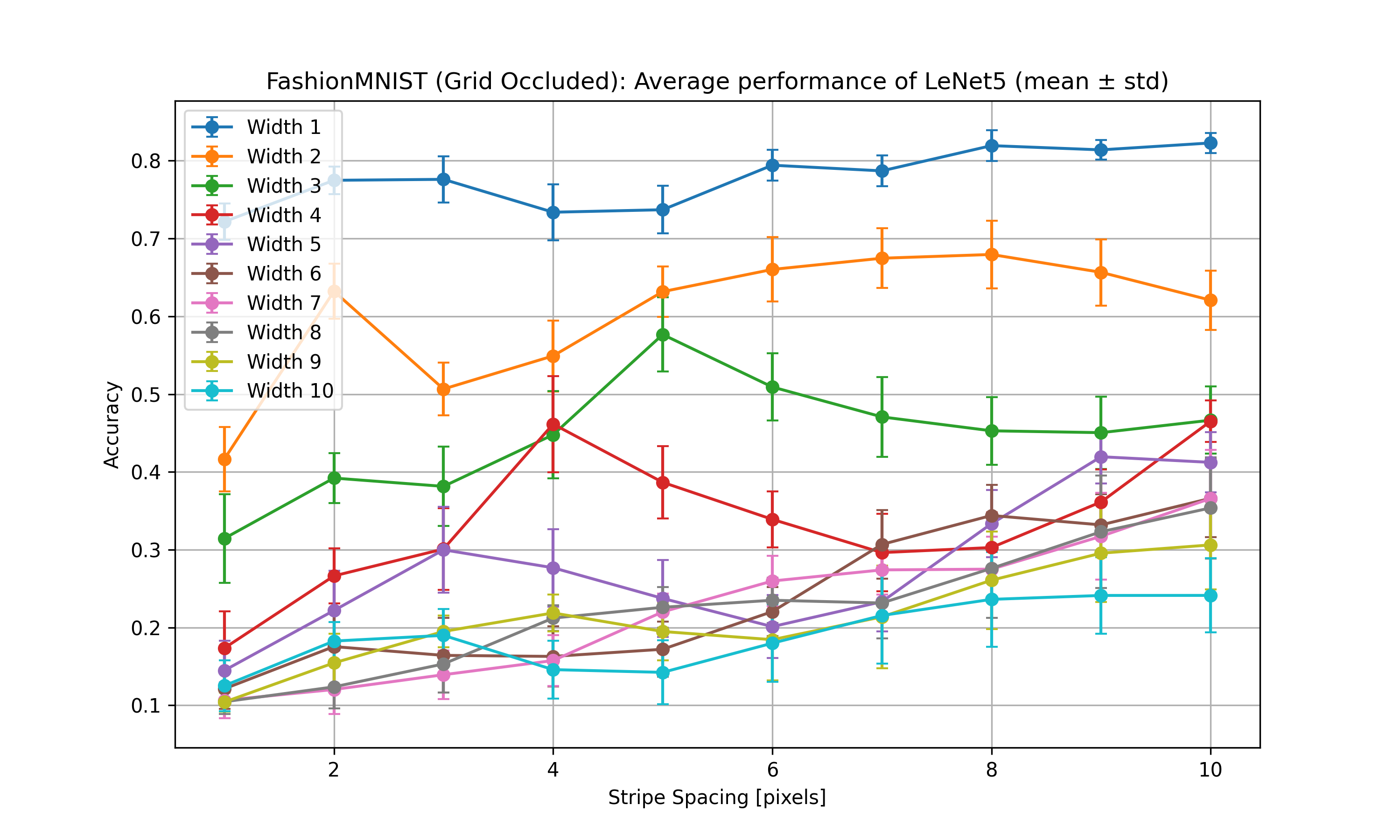}
    \caption*{(a) Vanilla LeNet5}  
  \end{minipage}%
  \begin{minipage}{0.5\textwidth}
    \centering
    \includegraphics[width=\linewidth]{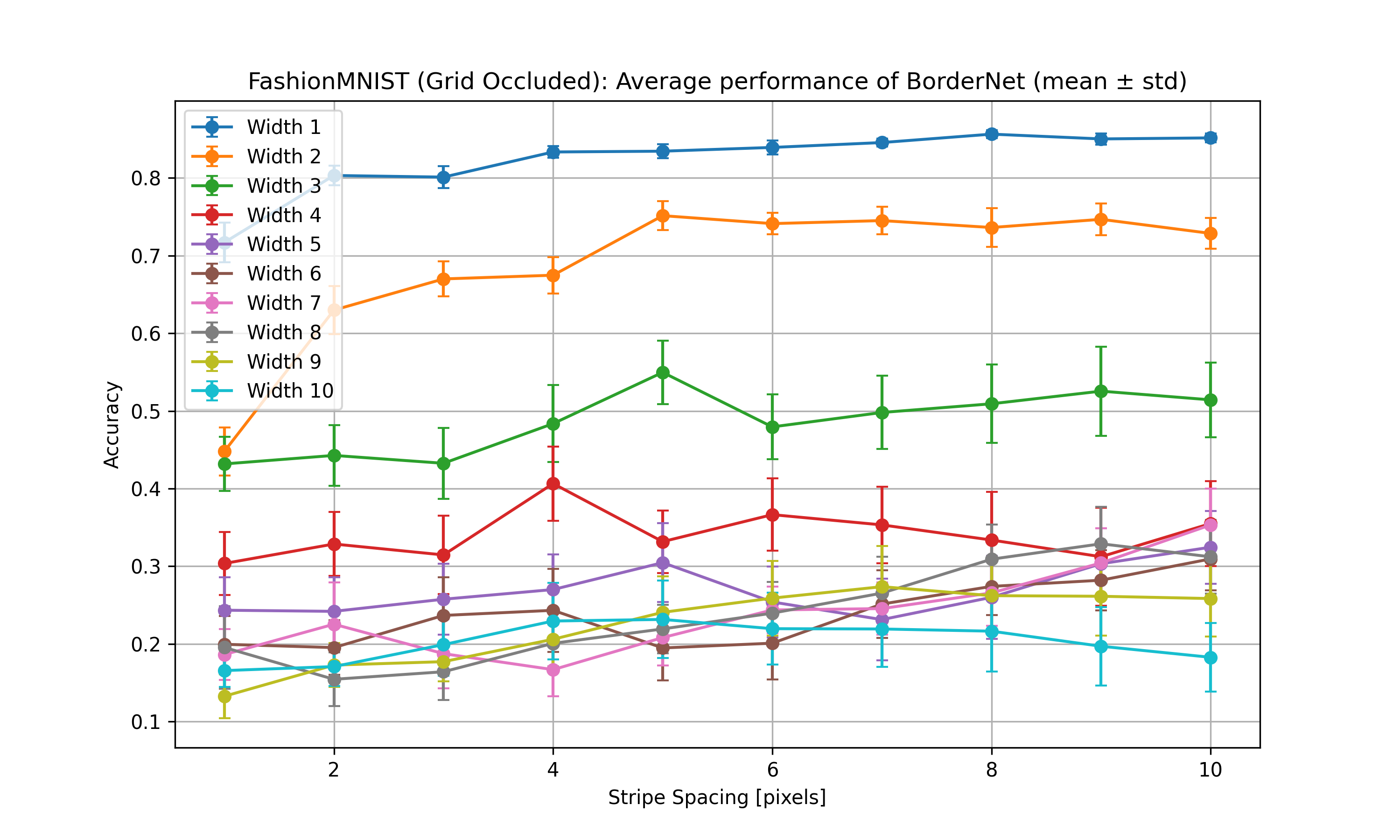}
    \caption*{(b) BorderNet} 
  \end{minipage}
  \caption{Accuracies with FashionMNIST occlusion grid spacing and width for the Vanilla LeNet5 and BorderNet.}  
  \label{fig:FashionMNIST_grid_perf}
\end{figure}

\vspace*{-10mm}

\begin{figure}[h]
  \centering
  \begin{minipage}{0.5\textwidth}
    \centering
    \includegraphics[width=\linewidth]{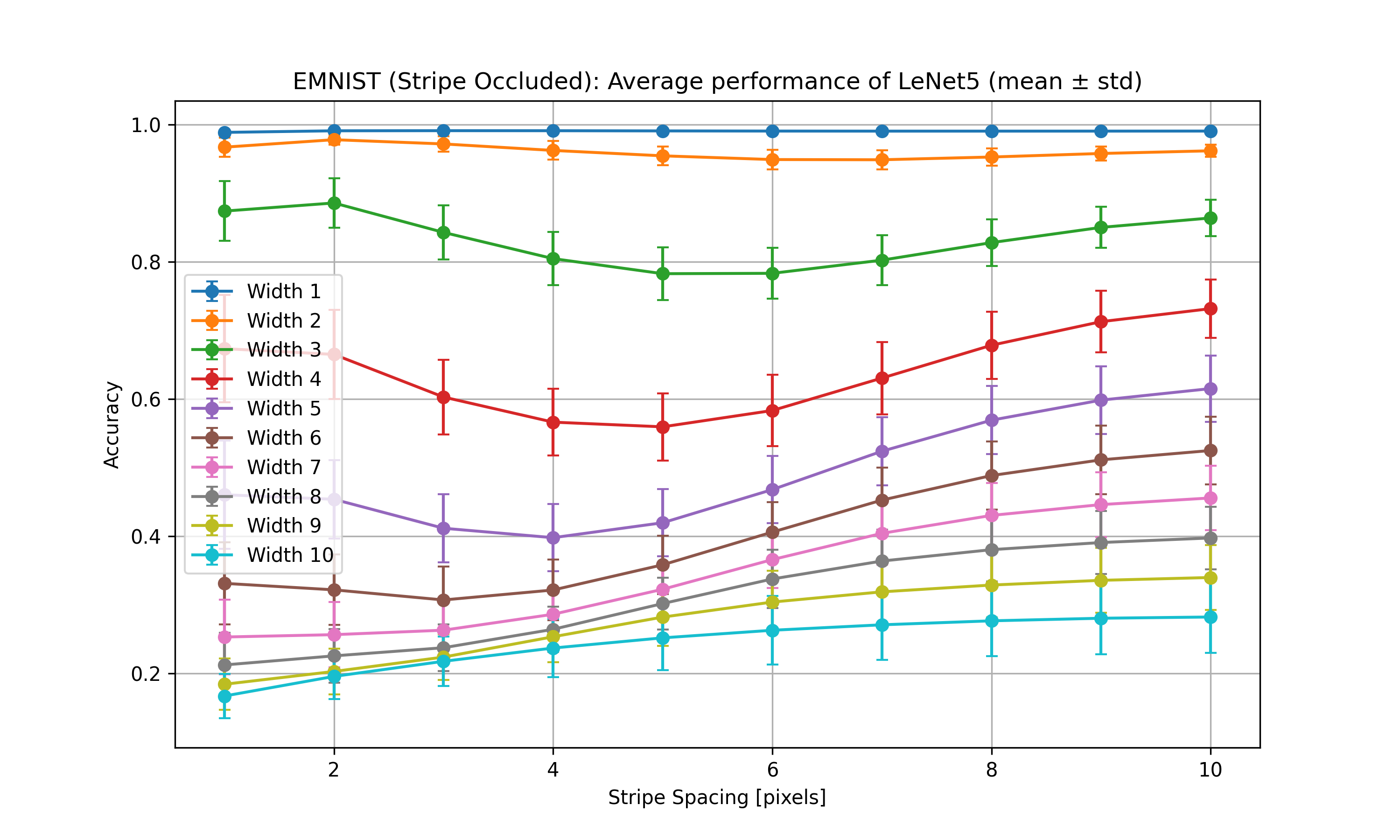}
    \caption*{(a) Vanilla LeNet5}  
  \end{minipage}%
  \begin{minipage}{0.5\textwidth}
    \centering
    \includegraphics[width=\linewidth]{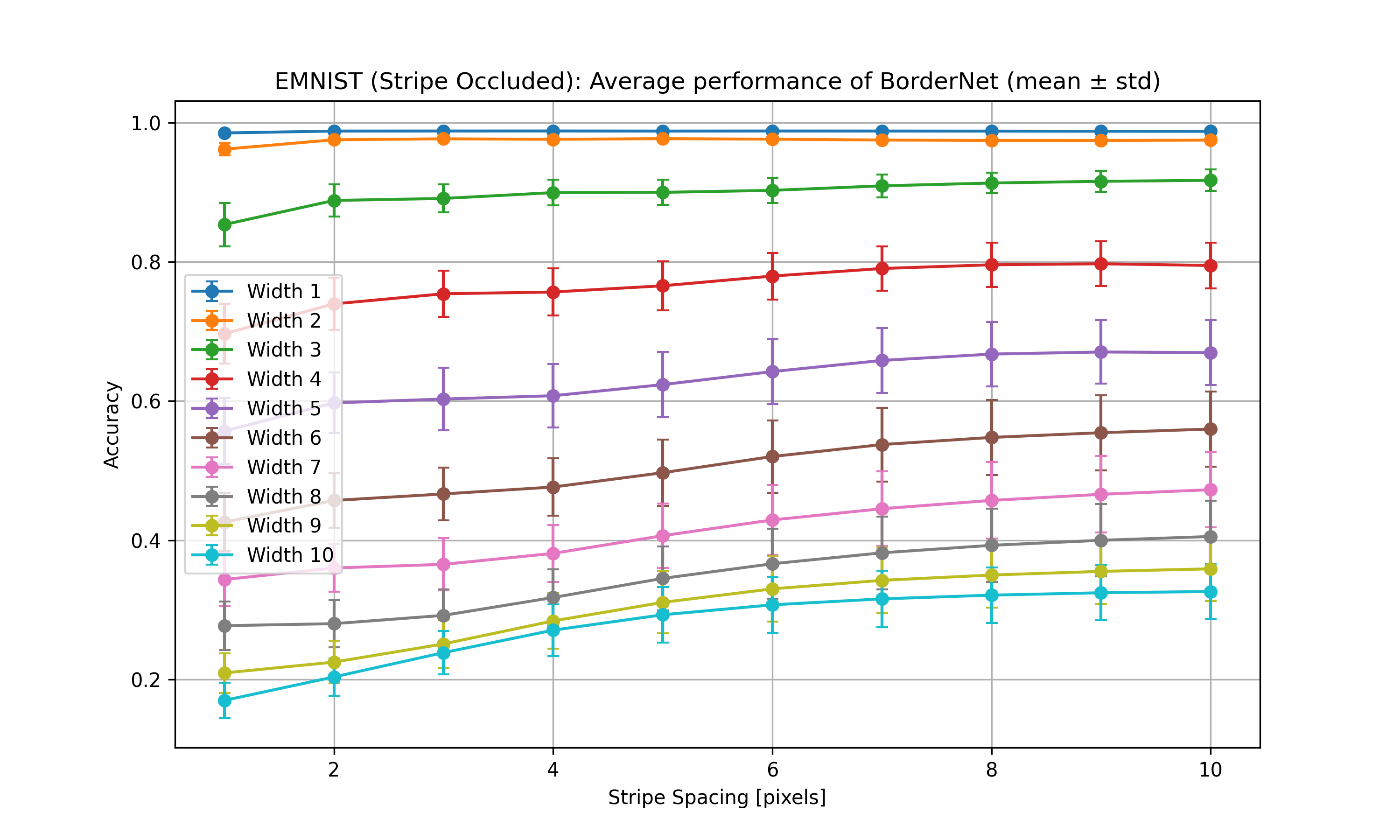}
    \caption*{(b) BorderNet} 
  \end{minipage}
  \caption{Accuracies with EMNIST occlusion stripe spacing and width for the Vanilla LeNet5 and BorderNet.}  
  \label{fig:EMNIST_stripe_perf}
\end{figure}

\vspace*{-10mm}

\begin{figure}[t]
  \centering
  \begin{minipage}{0.5\textwidth}
    \centering
    \includegraphics[width=\linewidth]{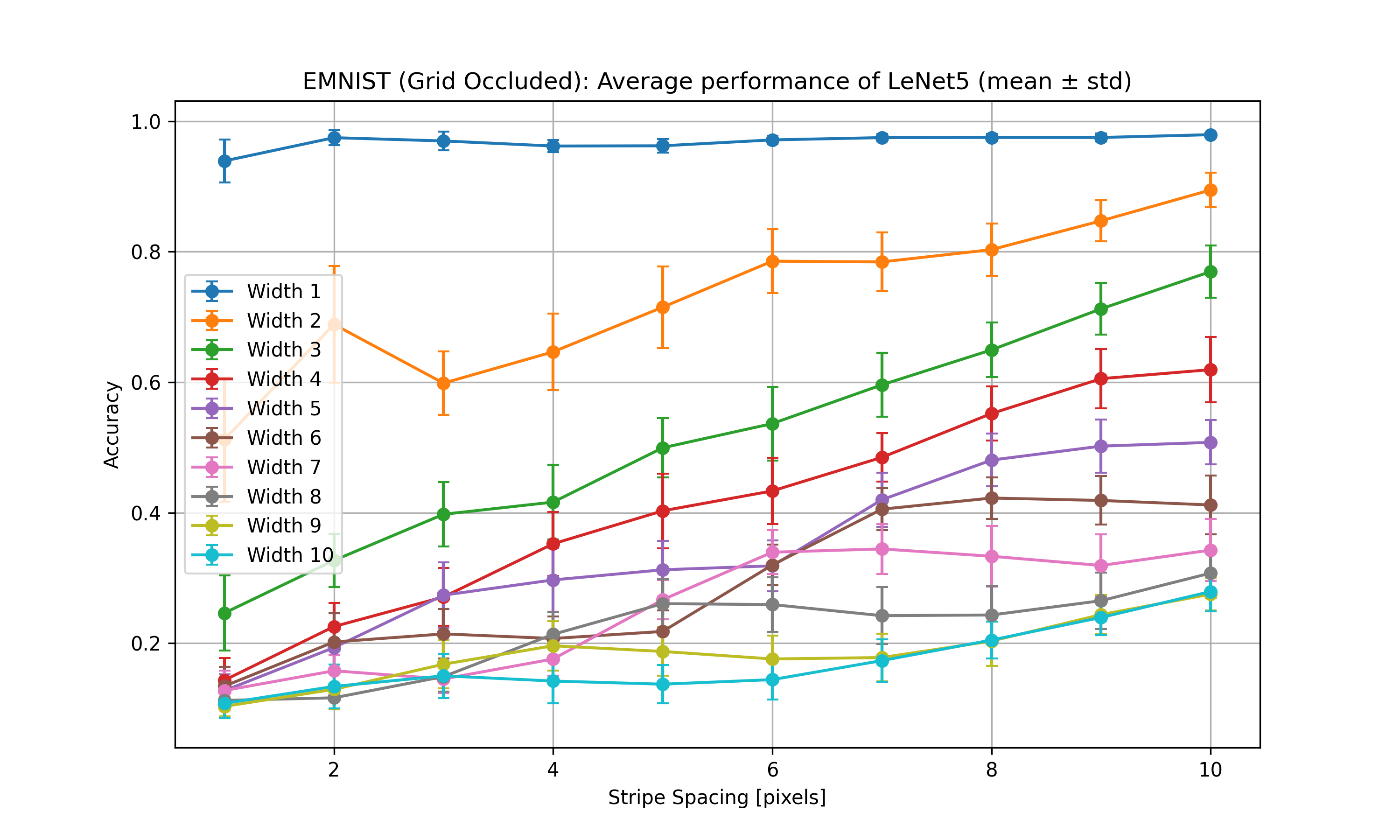}
    \caption*{(a) Vanilla LeNet5}  
  \end{minipage}%
  \begin{minipage}{0.5\textwidth}
    \centering
    \includegraphics[width=\linewidth]{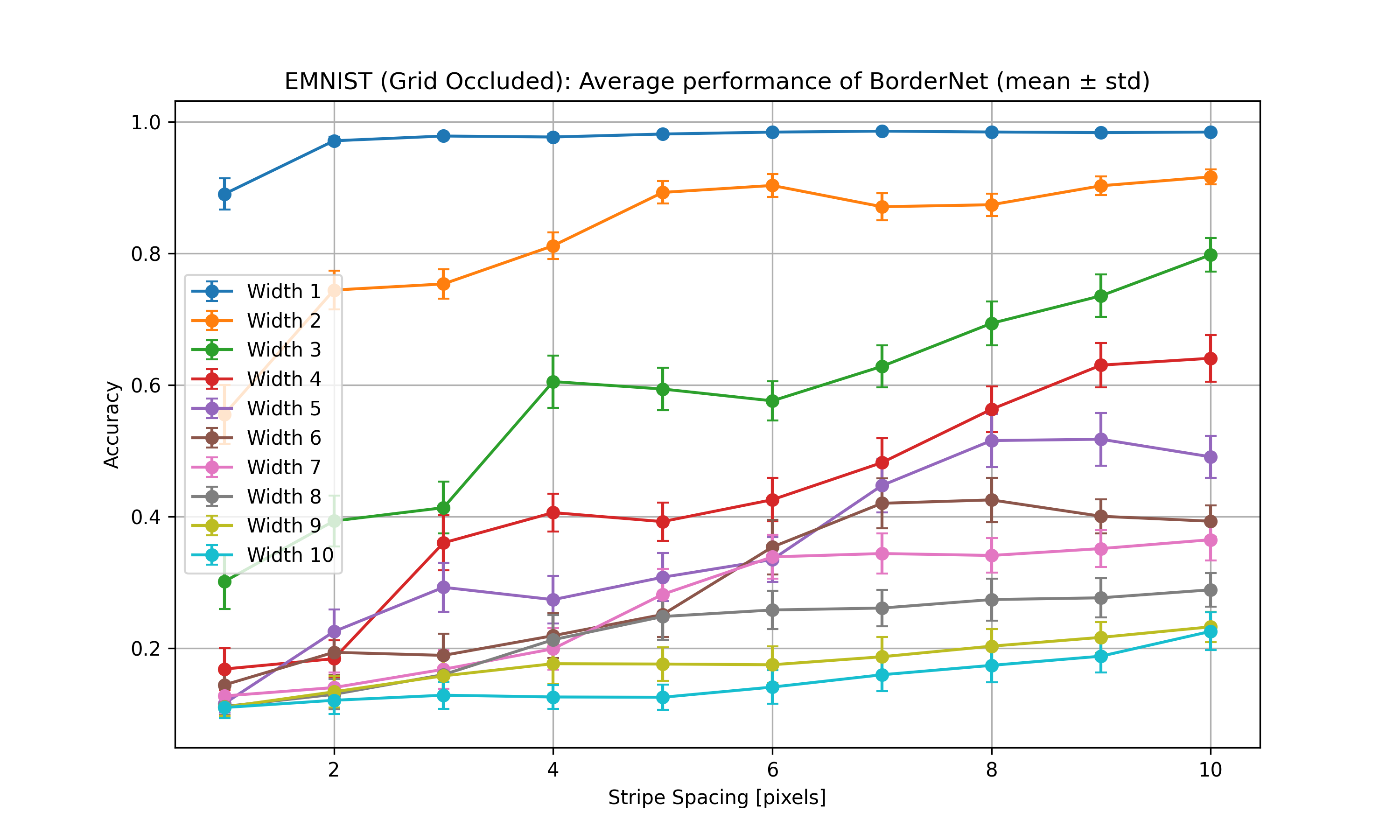}
    \caption*{(b) BorderNet} 
  \end{minipage}
  \caption{Accuracies with EMNIST occlusion grid spacing and width for the Vanilla LeNet5 and BorderNet.}  
  \label{fig:EMNIST_grid_perf}
\end{figure}

\end{document}